%% file: paper.tex
\documentclass[]{fairmeta}

\usepackage{tabularx}
\usepackage{textcomp}
\usepackage{url}
\usepackage{verbatim}
\usepackage{titlesec}
\usepackage{adjustbox}
\usepackage{multirow}
\usepackage{pifont}
\usepackage{tikz}
\usepackage{comment}
\usepackage{amsmath,amssymb} 
\usepackage{colortbl}  
\usepackage{booktabs} 
\usepackage{hyperref}
\usepackage{graphicx}    
\usepackage{subcaption} 
\definecolor{headerpurple}{HTML}{d8d2fc}
\RequirePackage{xspace}
\makeatletter
\DeclareRobustCommand\onedot{\futurelet\@let@token\@onedot}
\def\@onedot{\ifx\@let@token.\else.\null\fi\xspace}
\usepackage[most]{tcolorbox}
\usepackage{xcolor}
\usepackage{enumitem}
\usepackage{mdframed}
\definecolor{keywordcolor}{RGB}{178,34,34} 
\usepackage{siunitx}

\definecolor{tablepurple}{HTML}{E4DDF9}

\newcolumntype{C}{>{\centering\arraybackslash}X}

\newcommand{\huggingface}{\raisebox{-1.5pt}{\includegraphics[height=1.05em]{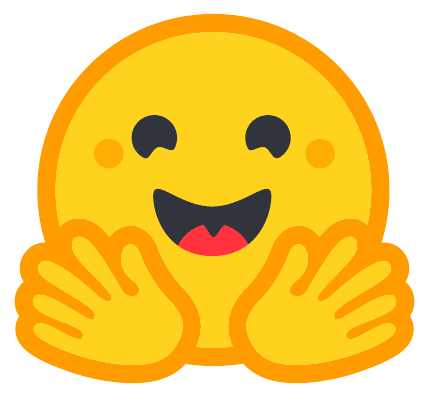}}\xspace}
\newcommand{\github}{\raisebox{-1.5pt}{\includegraphics[height=1.05em]{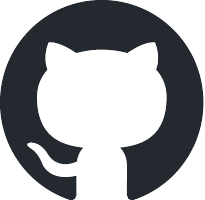}}\xspace}
\newcommand{\homepage}{\raisebox{-1.5pt}{\includegraphics[height=1.05em]{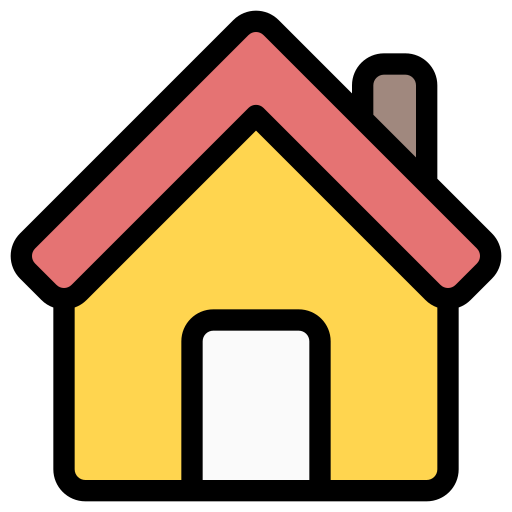}}\xspace}
\newcommand{\modelscope}{\raisebox{-1pt}{\includegraphics[height=0.95em]{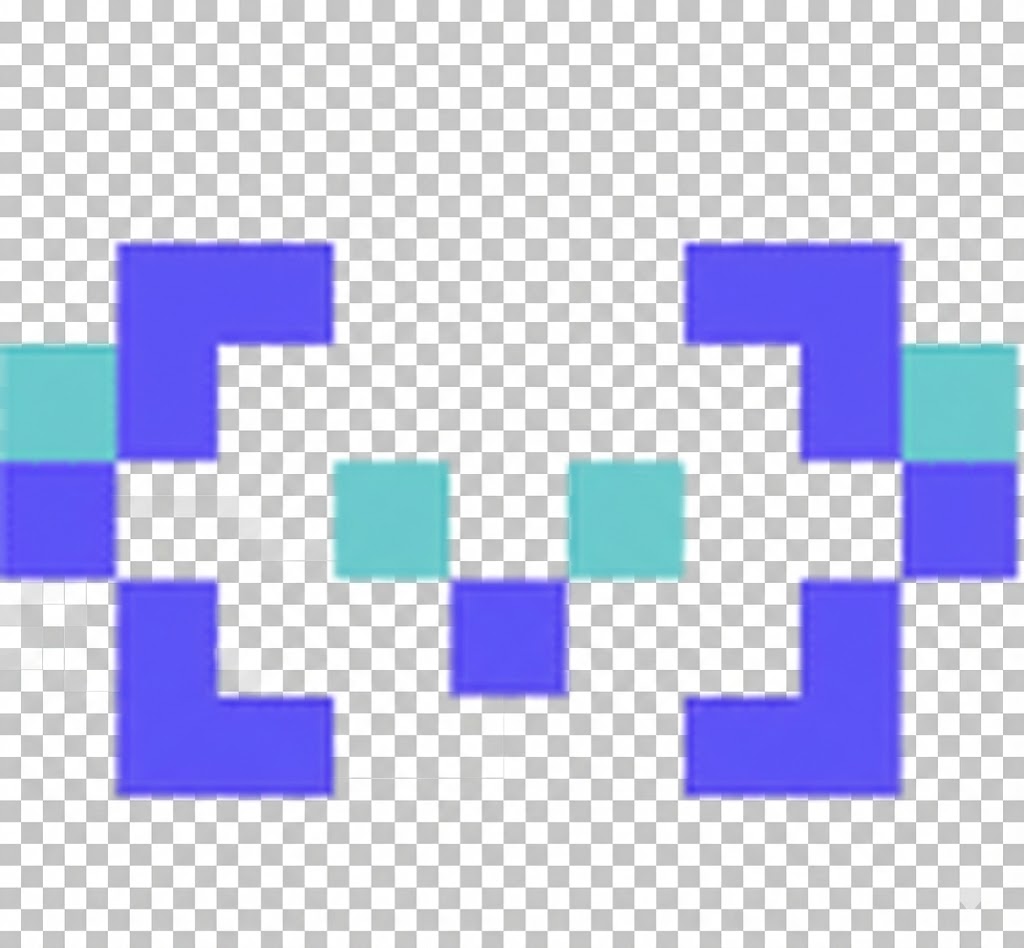}}\xspace}

\makeatother

\definecolor{adptorange}{RGB}{248, 205, 172}
\definecolor{cmpblue}{RGB}{189, 215, 238}

\definecolor{our_red}{RGB}{232,157,160}
\definecolor{our_blue}{RGB}{136,206,230}
\definecolor{our_orange}{RGB}{246,200,168}
\definecolor{our_green}{RGB}{178,211,164}

\definecolor{attn_code0}{RGB}{247,215,200}
\definecolor{attn_code1}{RGB}{238,169,139}
\definecolor{mlp_code0}{RGB}{204,201,221}
\definecolor{mlp_code1}{RGB}{102,95,153}

\definecolor{token_blue}{RGB}{84, 120, 140}

\usepackage{bbding}       
\usepackage{fontawesome}
\usepackage{xspace}

\usepackage{float}
\newcommand{\our}{{RynnWorld-Latent}\xspace}
\newcommand{\lam}{{RynnLAM}\xspace}

\newlength\savewidth

\newcolumntype{x}[1]{>{\centering\arraybackslash}p{#1pt}}
\newcolumntype{y}[1]{>{\raggedright\arraybackslash}p{#1pt}}
\newcolumntype{z}[1]{>{\raggedleft\arraybackslash}p{#1pt}}

\renewcommand{\paragraph}[1]{\vspace{1mm}\noindent\textbf{#1}}

\usepackage{algorithm}
\usepackage{listings}

\definecolor{codeblue}{rgb}{0.25, 0.5, 0.5}
\definecolor{codekw}{rgb}{0.35, 0.35, 0.75}
\lstdefinestyle{Pytorch}{
    language = Python,
    backgroundcolor = \color{white},
    basicstyle = \fontsize{9pt}{8pt}\selectfont\ttfamily\bfseries,
    columns = fullflexible,
    aboveskip=1pt,
    belowskip=1pt,
    breaklines = true,
    captionpos = b,
    commentstyle = \color{codeblue},
    keywordstyle = \color{codekw},
}

\definecolor{green}{HTML}{009000}
\definecolor{red}{HTML}{ea4335}

\title{\our: A Cross-Embodiment World Model Conditioned on Universal Latent Actions}

\author[* 1, 2, 3]{Haoyu Zhao}
\author[* 1]{Yinzhou Tang}
\author[* 1]{Zixuan Wang}
\author[1]{Minghao Zhu}
\author[1]{Dongchi Huang}
\author[1]{Xingyue Zhao}
\author[2, 3]{\\ Qin Zhao}
\author[1, 4]{Kehan Li}
\author[\dagger 1, 4]{Siteng Huang}
\author[1, 4]{Xin Li}
\author[1, 4]{Zhenyu Zeng}
\author[1, 4]{Mingxiu Chen}
\author[1, 4]{Deli Zhao}
\author[\dagger 2, 3]{Zhongyu Li}

\affiliation[1]{DAMO Academy, Alibaba Group,}
\affiliation[2]{Hong Kong Embodied AI Lab,}
\affiliation[3]{CUHK,}
\affiliation[4]{Hupan Lab,\\}

\contribution[*]{Core contributors}
\contribution[\dagger]{Corresponding author}

\abstract{
Learning controllable world dynamics from heterogeneous interaction videos is challenging because robot embodiments use incompatible action spaces, while human videos lack native control labels. We present \our, a cross-embodiment video world model that uses the 3D-grounded latent actions introduced by \lam as a shared conditioning interface.
This representation enables joint pretraining on human interaction videos, robot demonstrations, and simulation rollouts without requiring a common native action space. Built on a pretrained video transformer, \our combines global action tokens with an identity-initialized local pathway that aligns actions with future video latents. We pretrain on 23 datasets comprising approximately 553.5M frames and adapt the model to target robots through post-training. The framework supports two data-generation routes: digital teleoperation uses target robot commands to generate action-video demonstrations, while latent-action synthesis transfers source motion to an edited initial frame and pairs the resulting rollout with retargeted pseudo robot actions. Experiments show improved video prediction metrics on public benchmarks and two real-robot platforms, alongside qualitative action following when different action sequences share the same initial observation. Synthetic mid-training improves policy success rates on LIBERO and LIBERO-Plus, and both generation routes improve real-world policy success rates over real-demonstration-only training in the evaluated settings.
These results support universal latent actions as a scalable interface for learning controllable visual dynamics across heterogeneous embodiments.

\begin{center}
    \renewcommand{\arraystretch}{1.2}
    \begin{tabular}{ll}
        \homepage  & \url{https://alibaba-damo-academy.github.io/RynnWorld-Latent.github.io} \\
        \github  & \url{https://github.com/alibaba-damo-academy/RynnWorld-Latent}\\
        \huggingface & \url{https://huggingface.co/Alibaba-DAMO-Academy/RynnWorld-Latent} \\
        \modelscope  & \url{https://www.modelscope.cn/models/DAMO_Academy/RynnWorld-Latent}\\
    \end{tabular}
\end{center}

}

\date{\today}

\begin{document}
\thispagestyle{firstheader}
\maketitle
\pagestyle{empty}

\input{Sec/1-intro}

\input{Sec/2-related}

\input{Sec/3-method}

\input{Sec/4-exp}

\input{Sec/5-conclusion}

\newpage
\bibliographystyle{assets/plainnat}
\bibliography{paper}

\newpage
\beginappendix

\input{Sec/Appendix}

\end{document}

%% file: Sec/1-intro.tex
\section{Introduction}

World models, which predict future states conditioned on actions, have long been theorized as a cornerstone of generalist robot intelligence~\citep{sutton1991dyna,lecun2022path,richens2025general}. The recent progress of large-scale video generation has catalyzed a new class of video world models, in which future dynamics are represented directly in visual space~\citep{zhao2026rynnworld,zhao2026rynnworldteleop,agarwal2025cosmos,ali2025world,zhao2026towards,zhao2026high,zhao2025physsplat}. These models achieve impressive visual fidelity and provide a promising substrate for learning physical dynamics from large-scale video corpora. However, when applied to robotic manipulation, language-conditioned video generators face an interface mismatch: their primary control signal is a task-level description, while physical interaction is governed by continuous, frame-level causal interventions.

Language describes semantic intentions at a coarse temporal granularity. An instruction such as ``pick up the cup'' specifies a task outcome, but does not determine the precise trajectory by which the outcome is achieved. In contact-rich manipulation, small differences in timing, contact order, gripper state, or camera motion can lead to substantially different futures.
Without fine-grained motion or action conditioning, a text-conditioned model must represent multiple plausible trajectories for the same instruction, making precise control over manipulation dynamics challenging.
Moreover, natural language alone is not a convenient interface for specifying temporally precise interventions, such as closing the gripper earlier while maintaining the same preceding motion.
These limitations motivate an explicit action-conditioning interface for controllable video world modeling.

A natural solution is to condition world models on robot actions~\citep{gao2026dreamdojo}. However, directly using raw robot actions introduces a severe alignment bottleneck. Different embodiments operate in heterogeneous kinematic spaces, with different joint counts, coordinate conventions, control frequencies, and hardware limits. As a result, a world model conditioned on raw actions is difficult to scale across platforms. At the same time, large-scale human videos contain abundant interaction dynamics without robot-specific control labels, while simulation actions need not match a target robot's control space. What is needed is not a stronger action encoder for a particular robot, but a shared action interface that represents interaction-related visual change while abstracting away platform-specific control conventions.

We argue that latent actions provide such an interface. Instead of representing actions as robot commands, latent actions describe what changed between observations. Distilled from video in a self-supervised manner, continuous latent actions provide a shared description of visual transitions without requiring identical hardware control interfaces. Human hands, robot grippers, and simulated agents may differ drastically in their low-level control signals, yet their visual transitions can share motion and interaction patterns.
By encoding heterogeneous videos into a shared latent-action representation, a single world model can be pretrained on human interactions, robot demonstrations, and simulation rollouts under a unified action-conditioning objective, without requiring their native control spaces to be aligned.

In this work, we introduce \textbf{\our}, a cross-embodiment video world model built upon \textbf{\lam}~\citep{zhao2026rynnvla}, the 3D-grounded latent-action representation introduced in RynnVLA-Latent~\citep{zhao2026rynnvla}. While RynnVLA-Latent~\citep{zhao2026rynnvla} uses these descriptors as supervision targets for policy pretraining, our work investigates their complementary role as conditioning signals for learning controllable visual dynamics. \our predicts future video from an initial observation and a sequence of continuous latent actions, without requiring text conditioning or a common native robot action space. Technically, we combine global action tokens with an identity-initialized, frame-aligned local conditioning pathway within a pretrained video diffusion transformer. The local pathway exploits the temporal correspondence between latent-action extraction and video-latent compression, allowing each action descriptor to condition its corresponding future video latent.

A central idea of this work is that \our should not be viewed only as a future predictor, but as a \textbf{data-generation engine} for scalable robot learning. This leads to two complementary generation modes. First, after target-robot post-training, \our can be conditioned on target robot actions and support interactive digital teleoperation~\citep{zhao2026rynnworldteleop}: operator control signals are mapped to target actions and drive video generation, producing paired action-video data without requiring additional physical robot execution. Second, the latent-action-conditioned model enables data synthesis from existing videos. Given a source video, we extract its latent-action sequence, edit the first frame to replace the embodiment or environment, and generate a new rollout conditioned on the edited first frame and the original latent actions. First-frame editing specifies the new visual embodiment or environment, while the latent-action sequence guides the interaction dynamics. The resulting rollout has motion close to the source video, but appears under a new embodiment or environment. These generated rollouts can be paired with retargeted pseudo robot actions for downstream VLA training.

We evaluate \our from three complementary perspectives. First, we assess action-conditioned video prediction on public world-model benchmarks and held-out demonstrations from two real-robot platforms, measuring prediction fidelity after target-robot adaptation. Second, we investigate latent-action controllability through qualitative comparisons that vary the supplied latent-action sequences while holding the initial observation fixed, and illustrate cross-embodiment motion transfer through first-frame editing. Third, we examine the utility of world-model-generated data for downstream policy learning. Synthetic mid-training improves success rates on LIBERO and LIBERO-Plus, while both latent-action-driven synthesis and digital teleoperation improve observed closed-loop policy success on the evaluated real-world manipulation tasks. These evaluations jointly characterize prediction fidelity, qualitative action following, and downstream training utility.

Our main contributions are as follows:
\begin{itemize}
    \item We present \our, a text-free, latent-action-conditioned video world model,
    which uses frozen \lam descriptors as a shared conditioning interface for learning visual dynamics from human interaction videos, heterogeneous robot demonstrations, and simulation rollouts without aligning their native control spaces.

    \item We design an action-conditioning mechanism that combines global action tokens with locally identity-initialized, frame-aligned injection, associating transition descriptors with the corresponding future video latents.

    \item We enable latent-action-guided generation under edited robot embodiments and environments, and adapt the world model to native robot-action conditioning. The resulting latent-action synthesis and digital-teleoperation pipelines generate data that improves downstream policy learning in the evaluated settings.
    
\end{itemize}

%% file: Sec/2-related.tex
\section{Related Work}
\label{sec:related_work}

\subsection{World Models}
World models predict how environments change under actions and support planning and policy learning through simulated experience~\citep{finn2016physical,ebert2018foresight,hafner2019planet,hafner2020dreamer}. Advances in video generation have enabled large-scale pretraining of visual dynamics models and their adaptation to interactive simulation~\citep{agarwal2025cosmos,yang2024unisim,wu2024ivideogpt,rigter2024avid,huang2025vid2world}. Language-conditioned video and image generation have also been used to produce visual plans and synthetic demonstrations for robot control~\citep{du2023unipi,zhou2024robodreamer,black2024susie,jang2025dreamgen}. However, task-level instructions leave the timing and trajectory of individual movements largely unspecified. To simulate fine-grained manipulation, action-conditioned world models incorporate robot controls into future prediction~\citep{zhu2025irasim,guo2026ctrlworld,wang2025hma}. Such predictions support policy evaluation~\citep{li2025worldeval,quevedo2026worldgym,wang2026iws} and planning~\citep{ebert2018foresight,zhou2025dinowm,assran2025vjepa2,kim2026cosmospolicy}, while related approaches combine future prediction with action learning to improve robot policies~\citep{wu2024gr1,cheang2024gr2,zhu2025uwm,ye2026dreamzero,sun2026vlajepa}. 

Extending action-conditioned modeling across embodiments requires handling differences in action dimensions, coordinate conventions, and control semantics. Existing approaches address this through heterogeneous action modeling~\citep{wang2025hma} or geometric motion representations shared by humans and robots~\citep{wu2026oscar,zhao2026rynnworldteleop}. For pretraining on diverse robot and human videos, a further challenge is that many data sources lack corresponding control labels. More closely related to our setting, latent-action pretraining infers conditioning variables from observed transitions, allowing videos without native action labels to contribute to world-model learning~\citep{gao2025adaworld,garrido2026wild,gao2026dreamdojo,liu2026clapworld}. Building on this direction, \our uses continuous latent actions as a shared conditioning interface for cross-embodiment world-model pretraining, without requiring direct alignment of embodiment-specific control spaces.

\subsection{Latent Action Learning}
Learning across embodiments requires representations that connect different forms of observation and control. Prior work studies shared policies trained on heterogeneous robot data~\citep{o2024open,octo2024octo,kim2024openvla,wang2024hpt} and action-space alignment using recorded controls~\citep{zheng2025uniact,zhang2025ate}. For videos without corresponding robot actions, transferable skills and geometric motion cues provide intermediate representations for learning~\citep{xu2023xskill,kim2025uniskill,wen2024atm,bharadhwaj2024track2act,collins2025amplify}. Latent action learning instead infers compact action variables from transitions between observations~\citep{rybkin2019clasp,schmidt2024lapo}. These variables have been used as supervision for policy pretraining and transfer~\citep{ye2025lapa,chen2025moto,bu2025univla,chen2025villax,li2025latbot,tharwat2025lawm}, as well as conditioning inputs for predicting future states~\citep{menapace2021playable,bruce2024genie,alles2025latentworld,bi2025motus,garrido2026wild}.

Both discrete codes~\citep{bruce2024genie,ye2025lapa,chen2025moto} and continuous representations~\citep{gao2025adaworld,liang2026clam,yang2026como,garrido2026wild,gao2026dreamdojo} have been explored. Recent work further studies alignment with available action supervision~\citep{zhang2026claplatent,luo2026jala}, the separation of motion from appearance~\citep{yang2026como,xu2026motionfocused,zhang2026dila}, and geometric constraints on the learned action
space~\citep{yang2026lawm3d}. These distinctions matter because visual changes can include camera motion, background variation, and other effects unrelated to the agent's control~\citep{nikulin2025laom,zhang2025lamanalysis,garrido2026wild}.
Building on these studies, we use a pretrained and frozen encoder \lam~\citep{zhao2026rynnvla} to extract 3D-grounded continuous latent-action descriptors from heterogeneous visual transitions. Unlike methods that require native control annotations or explicit motion retargeting during world-model pretraining, our approach uses these descriptors as a shared conditioning interface for human, robot, and simulation videos. This enables joint learning of visual dynamics without requiring the original action parameterizations to be aligned across embodiments.

%% file: Sec/3-method.tex
\section{\our}
\label{sec:method}


\begin{figure}[t]
    \centering
    \includegraphics[width=\linewidth]{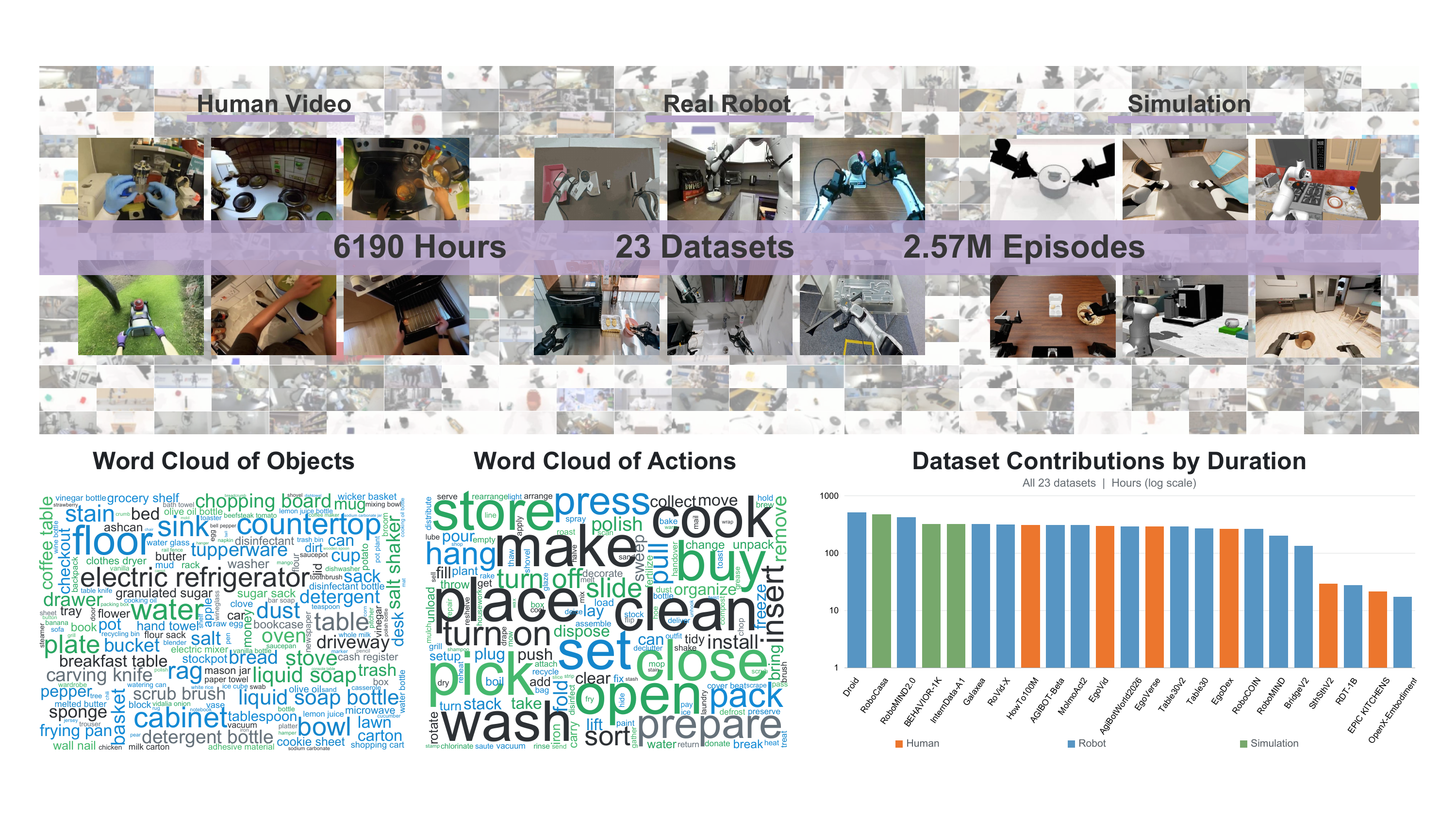}
    \caption{
    \textbf{Pretraining corpus overview.}
    The corpus contains 23 datasets, approximately 6,190 hours, and 2.57M episodes. The figure summarizes the distribution of human, real-robot, and simulation video, dataset contributions by duration, and object and action vocabulary.
    }
    \label{fig:Datasets}
\end{figure}


\begin{table*}[t]
  \centering
  \caption{Dataset-level statistics of the pretraining data.
  FPS denotes the frame rate declared in the training data metadata;
  comma-separated values indicate subsets with different frame rates. 
  Hours and frame counts are rounded to two decimal places; episode counts are integers. Frame rates describe metadata rather than a uniform sampling rate across all sources.}
  \label{tab:pretraining-data-statistics}
  \small
  \setlength{\tabcolsep}{6pt}
  \renewcommand{\arraystretch}{1.08}
  
  \sisetup{
      group-separator = {,},
      group-minimum-digits = 4,
      table-number-alignment = center,
      detect-weight = true,
      mode = text,
      round-mode = places,
      round-precision = 2
  }

  \begin{adjustbox}{max width=\linewidth}
    \begin{tabular}{@{}lc
          S[table-format=5.2]
          S[table-format=8.0, round-mode=none]
          S[table-format=5.2]@{}}
      \toprule
      \textbf{Dataset} & \textbf{FPS} & \textbf{Hours} & \textbf{Episodes} & {\textbf{Frames (M)}} \\
      \midrule
      
      \multicolumn{5}{@{}l}{\textit{Robot data}} \\
      \addlinespace[2pt] 
      
      DROID~\citep{khazatsky2024droid} & 15 & 518.55 & 82585 & 28.002 \\
      RoboMIND~\citep{wu2024robomind} & 20 & 200.21 & 42919 & 14.415 \\
      RoboMIND2.0~\citep{hou2025robomind} & 15 & 569.26 & 71226 & 30.740 \\
      Galaxea~\citep{jiang2025galaxea} & 15 & 320.49 & 17013 & 17.364 \\
      RoVid-X~\citep{deng2026rethinking} & 30 & 317.93 & 390614 & 34.336 \\
      AGIBOT-Beta~\citep{bu2025agibot} & 30 & 310.89 & 128955 & 33.576 \\
      AgiBotWorld2026~\citep{bu2025agibot} & 30 & 295.00 & 20704 & 31.860 \\
      MolmoAct2~\citep{fang2026molmoact2} & 30 & 308.62 & 32182 & 33.331 \\
      Table30~\citep{yakefu2025robochallenge} & 30 & 270.44 & 25585 & 29.207 \\
      Table30v2~\citep{yakefu2025robochallenge} & 30 & 290.92 & 32875 & 31.420 \\
      RoboCOIN~\citep{wu2025robocoin} & 30 & 262.95 & 92422 & 28.666 \\
      BridgeV2~\citep{walke2023bridgedata} & 5 & 135.25 & 13454 & 2.434 \\
      RDT-1B~\citep{liu2025rdt} & 30 & 27.69 & 6076 & 2.990 \\
      OpenX-Embodiment~\citep{o2024open} & 30 & 17.30 & 3550 & 0.819 \\
      
      \addlinespace[1pt]
      \rowcolor{gray!10} \textbf{Subtotal} & -- & \bfseries 3845.50 & \bfseries 960160 & \bfseries 319.16 \\

      \midrule
      \multicolumn{5}{@{}l}{\textit{Robot data (simulation)}} \\
      \addlinespace[2pt]
      
      RoboCasa~\citep{nasiriany2024robocasa} & 20 & 478.90 & 263919 & 34.481 \\
      BEHAVIOR-1K~\citep{li2023behavior} & 30 & 324.53 & 19971 & 35.050 \\
      InternData-A1~\citep{tian2026interndata} & 30 & 322.39 & 215443 & 34.818 \\
      
      \addlinespace[1pt]
      \rowcolor{gray!10} \textbf{Subtotal} & -- & \bfseries 1125.82 & \bfseries 499333 & \bfseries 104.35 \\

      \midrule
      \multicolumn{5}{@{}l}{\textit{Human interaction video}} \\
      \addlinespace[2pt]
      
      HowTo100M~\citep{miech2019howto100m} & 30 & 311.77 & 423975 & 32.036 \\
      EgoVid~\citep{wang2026egovid} & 30 & 299.00 & 397849 & 32.292 \\
      EgoVerse~\citep{punamiya2026egoverse} & 30 & 291.57 & 50212 & 31.489 \\
      EgoDex~\citep{hoque2025egodex} & 30 & 265.11 & 192175 & 28.632 \\
      SthSthV2~\citep{goyal2017something} & 12 & 29.71 & 26747 & 1.283 \\
      EPIC\_KITCHENS~\citep{damen2020epic} & 50 & 21.51 & 16905 & 4.262 \\
      
      \addlinespace[1pt]
      \rowcolor{gray!10} \textbf{Subtotal} & -- & \bfseries 1218.67 & \bfseries 1107863 & \bfseries 129.99 \\
      
      \midrule
      \addlinespace[1pt]
      \rowcolor{tablepurple} \textbf{Total (23 datasets)} & -- & \bfseries 6189.99 & \bfseries 2567356 & \bfseries 553.50 \\
      \bottomrule
    \end{tabular}
  \end{adjustbox}
\end{table*}


\subsection{Overview}
\our is a cross-embodiment latent-action-conditioned video world model that can further serve as a data-generation engine for robot learning. It predicts future video latents from an initial latent frame and a sequence of continuous latent actions, without relying on text conditioning. The pretraining corpus is summarized in Tab.~\ref{tab:pretraining-data-statistics} and Fig.~\ref{fig:Datasets}. 
The framework consists of three stages: 
(i) cross-embodiment pretraining, where human egocentric videos, robot teleoperation data, and simulation rollouts are relabeled into a shared latent-action space; 
(ii) target-robot post-training, which adapts \our to a target embodiment and introduces a robot-action conditioning interface;
and (iii) data generation, where the post-trained model supports interactive digital teleoperation when conditioned on target robot actions, and enables latent-action-driven data synthesis when conditioned on transferred latent actions.

We index video latent frames from $0$ to $L{-}1$, with the initial latent frame $x_0=c$ held fixed. Formally, given this conditioning frame and latent actions $A=\{\mathbf{z}_j\}_{j=1}^{J}$, \our learns
\begin{equation}
    p_\theta\!\left(x_{1:L-1} \;\middle|\; c,\, A\right).
\end{equation}
The latent-action interface is used for source-motion transfer and data synthesis; target-robot adaptation introduces robot-action conditioning for prediction and digital teleoperation.


\subsection{Latent-Action Interface}
Raw robot actions are embodiment-bound: joint counts, control modes, and coordinate conventions differ across platforms. We instead use continuous latent actions as a unified interface. For $j \in \{1,\ldots,J\}$, given frames $v_{4(j-1)}$ and $v_{4j}$, the frozen \lam~\citep{zhao2026rynnvla} produces
\begin{equation}
    \mathbf{z}_j
    =
    f_\phi(v_{4(j-1)}, v_{4j})
    =
    \big[\mathbf{z}_j^{m};\mathbf{z}_j^{a};\mathbf{z}_j^{c}\big]
    \in \mathbb{R}^{608}.
\end{equation}
Here $\mathbf{z}_j^{m}\in\mathbb{R}^{512}$ contains motion-token features, $\mathbf{z}_j^{a}\in\mathbb{R}^{64}$ is the global action latent, and $\mathbf{z}_j^{c}\in\mathbb{R}^{32}$ is the camera component. We group $[\mathbf{z}_j^{m};\mathbf{z}_j^{a}]$ into a 576-dimensional manipulation-related descriptor. Each action dimension is quantile-normalized to $[-1,1]$ using the $1$st and $99$th percentiles estimated from an 8,000-episode sample, and then clipped.

A full training chunk contains $T{=}81$ frames. A causal video VAE with $4{\times}16{\times}16$ compression~\citep{wang2025wan} maps it to $L{=}21$ latent frames. Since the LAM stride equals the VAE temporal compression factor, each latent action is index-aligned with one latent-frame transition, giving $J{=}L{-}1{=}20$.
This correspondence motivates our frame-aligned conditioning design in Sec.~\ref{sec:lac-wm}, allowing each action descriptor to condition its associated future video latent. 
For shorter episodes, we retain variable-length chunks with $T \in \{33,37,\dots,81\}$, corresponding to $J \in [8,20]$.


\subsection{Latent-Action-Conditioned World Model}    \label{sec:lac-wm}
\our builds on the Cosmos3 mixture-of-transformers video foundation model~\citep{agarwal2026cosmos}, which denoises video latents with rectified flow~\citep{liu2022flow}. The backbone does not use adaptive layer normalization, so action conditioning must enter through attention tokens or explicit feature modulation.

We encode each latent action with two linear projections and a shared residual MLP:
\begin{equation}
    h_j =
    W^{\mathrm{act}}\,[\mathbf{z}_j^{m}; \mathbf{z}_j^{a}]
    +
    W^{\mathrm{cam}}\,\mathbf{z}_j^{c},
    \qquad
    \tau_j = h_j + \mathrm{MLP}(h_j).
\end{equation}
The MLP output projection is zero-initialized. All corpora use a single shared domain slot, keeping the encoder embodiment-agnostic. The action tokens $\tau_j$ participate in joint self-attention, providing a global conditioning pathway.

Before the first transformer block, we additionally apply a local, frame-aligned pathway to the vision tokens of future latent frame $j$:
\begin{equation}
    q
    \leftarrow
    \bigl(q \odot (1+\gamma_j) + \beta_j\bigr)
    +
    g \odot \tau_j,
    \qquad
    q \in \mathcal{V}_{j}.
\end{equation}
Here $\mathcal{V}_j$ denotes the set of vision tokens belonging to future latent frame $j$, so the first action conditions frame $1$ and the last action conditions frame $L{-}1$. A single zero-initialized linear head maps $\tau_j$ to the modulation parameters $(\gamma_j,\beta_j)$, and $g$ is a zero-initialized per-channel gate. The conditioning frame $c$ is not modulated, and the local pathway is applied only under exact index alignment. Thus the two added vision-facing pathways are identity maps at initialization. This follows the zero-init convention of control-augmented diffusion models and makes the local action-dependent correction initially zero. Because the re-initialized action tokens also participate in joint attention, local identity initialization does not imply that the complete conditioned model is identical to the pretrained generator. The action-token stream itself is re-initialized because the pretrained action encoder is shape-incompatible with the 608-dimensional latent action space; re-initialized action-interface modules are trained with a $5\times$ learning-rate multiplier.


\subsection{Training Objective}
When present, actions are supplied as clean conditioning, while only future video latents are denoised. Let $X^{\mathrm{data}}=(x_1,\ldots,x_{L-1})$ denote the clean future latent sequence and $\epsilon$ Gaussian noise of the same shape. We follow the released rectified-flow parameterization, where $t{=}1$ corresponds to pure noise and $t{=}0$ corresponds to clean data:
\begin{equation}
    X_t = t\,\epsilon + (1-t)\,X^{\mathrm{data}},
    \qquad
    v_t = \epsilon - X^{\mathrm{data}}.
\end{equation}
The training objective is
\begin{equation}
    \mathcal{L}
    =
    \mathbb{E}_{X^{\mathrm{data}},c,A,\epsilon,t}
    \left\|
    u_\theta(X_t, t, c, A) - v_t
    \right\|^2.
\end{equation}
Only the first latent frame $c$ is kept clean as conditioning. This training objective is consistent with inference, where latent actions are also clean and only the video stream is denoised. Training timesteps are sampled from a logit-normal distribution with resolution shift $5$ for the 480p tier.

To support classifier-free guidance, we randomly drop action conditioning during training. With probability $0.1$, we drop the entire action sequence. Independently, with probability $0.1$ we drop the manipulation-related 576-dimensional slice, and with probability $0.1$ we drop the 32-dimensional camera slice. These branches support joint and slice-specific guidance, which we apply at inference with a guidance scale of $1.15$ (Sec.~\ref{sec:experiments}). During generation, denoising proceeds from $t{=}1$ toward $t{=}0$, while the initial frame and supplied action inputs remain fixed.


\subsection{World-Model Data Engine}

\begin{figure}[htbp]
    \centering
    \includegraphics[width=\linewidth]{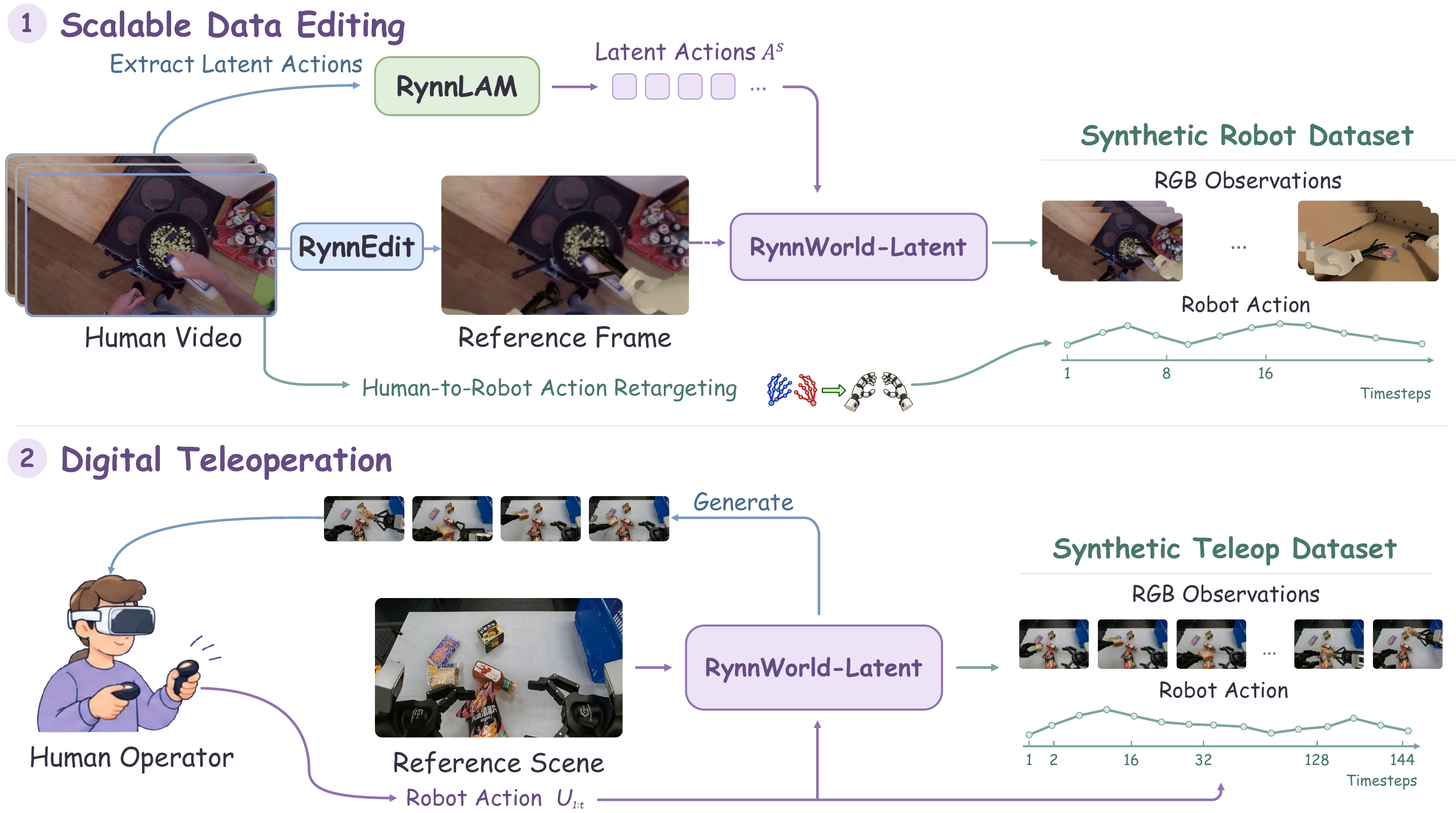}
    \caption{
    \textbf{Overview of the world-model data engine.}
    \textbf{Top: Scalable data editing.}
        \lam extracts latent actions from a source human video, while its first frame is edited to depict the target robot. Conditioned on the edited frame and the original latent actions, \our generates robot video rollouts. The generated observations are paired with pseudo-actions obtained through motion retargeting and inverse kinematics.
        \textbf{Bottom: Digital teleoperation.}
        Human control signals are retargeted to robot actions, which condition \our together with a reference scene. Generated frames provide visual feedback to the operator, while the recorded actions and generated observations form synthetic demonstrations without physical robot execution.
    }
    \label{fig:data_engine}
\end{figure}

To generate trajectories for policy learning, we combine the world model with interfaces for capturing interaction intent and producing action-video episodes. In our design, \our provides action-conditioned visual dynamics, while the surrounding pipeline provides target-robot adaptation, action retargeting, temporal alignment, and filtering. Compared with asset-based simulation, this generative pipeline can reduce the need for explicit 3D assets and manual system identification: scenes are specified implicitly by a conditioning image, visual outputs are produced in a learned video distribution, and interaction priors are drawn from large-scale video pretraining. We instantiate two complementary data-generation pipelines: interactive digital teleoperation and latent-action data synthesis.

\noindent
\textbf{Target-robot adaptation.}
Before using \our for data generation, we post-train it on target robot demonstrations. Our real-robot corpus is shared with RynnVLA-Latent~\citep{zhao2026rynnvla} and contains 3,850 collected episodes across six tasks (Appendix~\ref{app:real-data}). For each task, $5\%$ of the episodes are randomly held out for testing and the remaining $95\%$ form the target-robot post-training split; the 3,850-episode count refers to the full collected corpus. The target data is processed with the same chunking protocol and latent-action labeling pipeline. This stage adapts the model to the target embodiment's visual appearance, camera configuration, and interaction distribution. Robot-action conditioning is introduced through a separate target-specific encoder. When target robot actions are available, we introduce a lightweight action encoder $E_\psi$ that maps a target robot action $u_j$ into an action token:
\begin{equation}
    \tau_j^{u} = E_\psi(u_j).
\end{equation}
The token $\tau_j^{u}$ is injected through the same conditioning pathway as the latent-action token $\tau_j$. We train this interface with the same rectified-flow video objective, replacing latent-action tokens with robot-action tokens. Here $u_j$ denotes the robot-action input synchronized with the interval represented by latent action $\mathbf{z}_j$; its parameterization depends on the target robot. The encoder provides an embodiment-specific control interface without requiring commands to share \lam's descriptor dimensions.
After this adaptation, \our can be used not only as a latent-action-conditioned world model, but also as an action-conditioned world model for interactive generation.

\begin{figure}[t]
    \centering
    \includegraphics[width=0.95\linewidth]{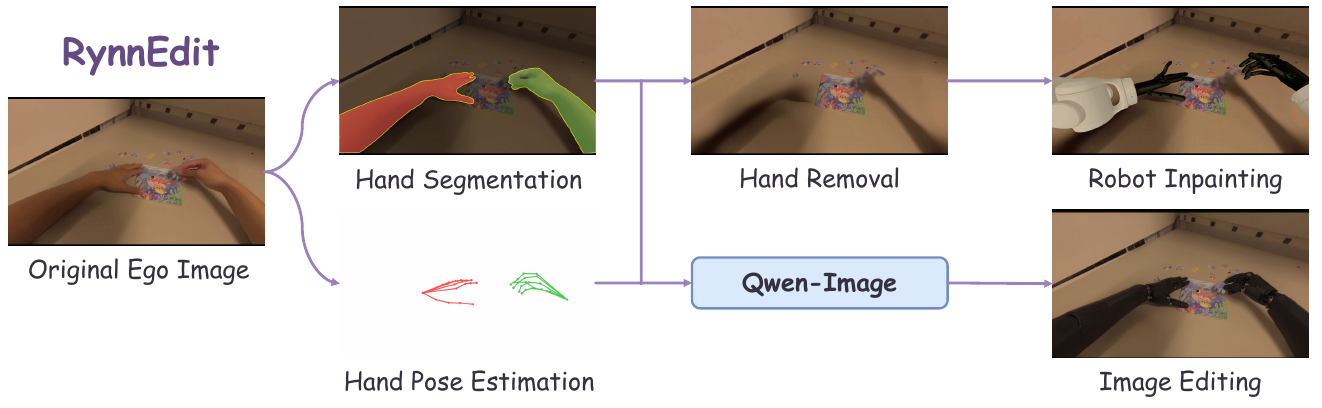}
    \caption{
    \textbf{RynnEdit.}
    Two routes construct a robot-centric reference frame: explicit hand removal, retargeting, and robot rendering; or direct replacement of human hands with robotic counterparts using an image-editing model. The resulting first frame conditions subsequent world-model generation.
    }
    \label{fig:Ego2Robot}
\end{figure}

\noindent 
\textbf{RynnEdit: Robot-centric reference-frame construction pipeline.} The data engine requires a reference scene frame $I_{\mathrm{ref}}$ compatible with the target robot's appearance and workspace. However, collecting real-world robot data in new environments is costly. We therefore synthesize scenes featuring the target robot from large-scale egocentric human hand videos. Given an egocentric interaction video, we construct a robot-centric initial observation by replacing human hands with the target robot while preserving the background and initial interaction configuration. The resulting first frame conditions the data engine; the source latent-action sequence supplies the subsequent motion.
Our pipeline shown in Fig.~\ref{fig:Ego2Robot} consists of egocentric data preprocessing followed by two alternative routes for generating robot-centric initial scenes: robot inpainting like Ego2Robot~\citep{wang2026ego2robot} and image editing using an end-to-end model. During preprocessing, we verify the integrity of the input videos and motion data, standardize their formats, and extract egocentric frames and human hand poses. We temporally align the video and motion sequences using zero-order hold or interpolation. We also use SAM3~\citep{carion2026sam}, prompted with human hand keypoints and text descriptions, to segment the hands and combine the resulting masks. In the robot inpainting route, we perform video inpainting within the hand masks to reconstruct the background after removing the human hands. We retarget the extracted human hand motion to the robotic hand and estimate the robot base pose and joint configurations through motion post-processing and inverse kinematics with base-pose search in MuJoCo~\citep{todorov2012mujoco}. We then render the robot in the estimated poses and composite the rendered images onto the inpainted background to obtain robotic manipulation observations corresponding to the original human demonstrations. In the image editing route, we instead provide the original egocentric image and its hand mask to an end-to-end image-editing model, like Qwen-Image-Edit~\citep{wu2025qwenimagetechnicalreport}, which replaces the human hands with robotic counterparts to generate an initial scene. The two routes thus produce synthetic robot observations through explicit retargeting and rendering or direct image editing, respectively. Finally, we apply quality gates to filter the generated observations and retain accepted samples as initial scenes for subsequent world-model-based data generation.

\noindent 
\textbf{Interactive digital teleoperation.}
The teleoperation pipeline converts operator motion into target robot actions and uses \our as the visual rollout engine (Fig.~\ref{fig:data_engine}). Given an initial conditioning frame $I_{\mathrm{ref}}$ and a stream of operator control signals, we retarget the operator motion to the target robot action space, producing a time-stamped action sequence. In chunk-based generation, the robot commands are synchronized with the video timeline and encoded through $E_\psi$. \our conditions on the current frame and the corresponding sequence of robot-action tokens to generate the future video latents. The last generated frame can then be used as the conditioning frame for the next chunk, enabling long-horizon interactive rollouts. Since the system records both the retargeted actions and the generated frames, interactive digital teleoperation directly produces paired action-video episodes without requiring physical robot execution.

\noindent 
\textbf{Latent-action data synthesis.}
The second pipeline synthesizes new robot data from existing videos using latent-action transfer. Given a source video $V^{s}$, we first extract its latent-action sequence $A^{s}$ using the frozen LAM. We then edit the first frame $I^{s}_{0}$ to obtain a new reference frame $\tilde{I}_{0}$ that replaces the embodiment or environment while preserving the scene layout, object arrangement, and camera viewpoint as much as possible. The edited frame is encoded as the conditioning latent $\tilde{c}$, and \our generates a rollout conditioned on this edited frame and the original latent-action sequence:
\begin{equation}
    \tilde{V}
    \sim
    p_\theta\!\left(\cdot \;\middle|\; \tilde{c},\, A^{s}\right).
\end{equation}
In this pipeline, first-frame editing specifies the target visual embodiment or environment, while the transferred latent-action sequence guides the generated interaction dynamics. The edited frame implicitly specifies the initial scene configuration without requiring explicit numerical object or robot states. The resulting rollout is encouraged to exhibit motion patterns related to the source interaction under the edited visual conditions. This transfer is approximate, since changes in robot geometry, workspace constraints, contact configurations, or viewpoint may affect whether the source motion remains physically feasible. The world model itself remains text-free: any prompts used by the external image editor only construct the reference frame $\tilde{I}_{0}$ and do not condition the subsequent video generation.

\noindent
\textbf{Action relabeling, synchronization, and filtering.}
For interactive digital teleoperation, target robot actions can be logged directly. For latent-action data synthesis, robot actions can be obtained by relabeling the source motion. When the source video already has robot actions compatible with the target embodiment, those actions can be reused after temporal alignment. Otherwise, we derive pseudo robot actions by estimating the relevant source motion, retargeting it to the target robot workspace, and applying inverse kinematics; implementation details are provided in Appendix~\ref{app:ik-relabeling}. The pseudo actions are temporally aligned with the generated video in frame-index space. We emphasize that temporal synchronization does not by itself establish physical consistency: the actions come from retargeting or inverse kinematics, while the observations are produced by an independently trained generator. We therefore treat the derived actions as pseudo-labels and filter episodes based on joint limits, motion smoothness, visual quality, temporal consistency, and plausibility of object interactions. The resulting synthetic episodes can be combined with real robot data for downstream VLA training.

%% file: Sec/4-exp.tex
\section{Experiments}
\label{sec:experiments}

\subsection{Experimental Setup}

\noindent
\textbf{World-model implementation.}
\our is built on Cosmos3-Edge~\citep{agarwal2026cosmos}, a mixture-of-transformers video foundation model. We train for $20$K optimizer steps with a global batch size of $512$, using AdamW with $(\beta_1,\beta_2)=(0.9,0.99)$, weight decay $0.05$, gradient-norm clipping $1.0$, a peak learning rate of $2.26\times10^{-4}$, $500$ warmup steps, and linear decay to $10\%$ of the peak. The newly introduced action-facing modules are re-initialized for the $608$-dimensional latent-action space and use a $5\times$ learning-rate multiplier. Training uses bf16 with FP32 master weights and optimizer states, and a power-function EMA with $s=0.1$. Parallelism uses hybrid-sharded FSDP with shard degree $8$ per node and replication degree $4$ across nodes. All videos are trained in 480p aspect-ratio buckets ($832\times480$, $640\times640$, $736\times544$, $544\times736$, and $480\times832$).


\noindent
\textbf{Policy backbones.}
For policy experiments, we use $\pi_{0}$~\citep{black2024pi_0} and $\pi_{0.5}$~\citep{intelligence2025pi05} as base VLA models. For all settings, we use the same downstream training protocol unless otherwise specified. For synthetic pretraining experiments, we use the term \textbf{mid-training} to denote continued pretraining on synthetic data, and \textbf{post-training} to denote subsequent fine-tuning on target task or robot data.

\begin{figure}
  \centering
    \includegraphics[width=0.9\linewidth]{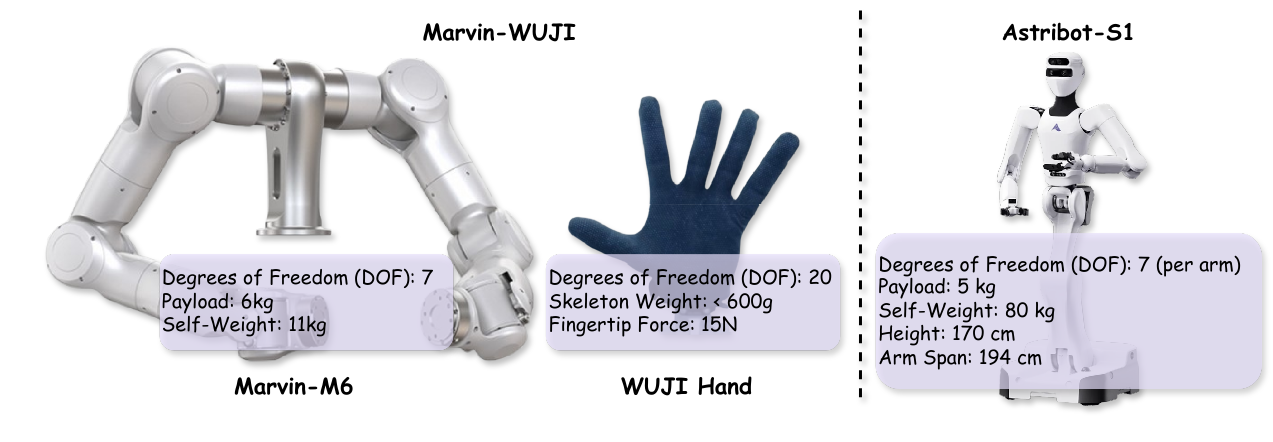}
  \caption{\textbf{Standardized Experimental Platforms.} Two dual-arm embodiments used for real-world evaluation: the stationary \textbf{Marvin-WUJI} (dual 7-DOF arms with 20-DOF dexterous hands) and the mobile \textbf{Astribot-S1} (dual 7-DOF arms with parallel grippers on a wheeled base).}
  \label{fig:vis-robot}
\end{figure}

\noindent
\textbf{Real-robot system setup.}
Our real-world experiments are conducted on two distinct robotic platforms to evaluate world-model adaptation and downstream policy learning, as shown in Fig.~\ref{fig:vis-robot}. For downstream policy deployment on both platforms, the inference frequency is set at 50 Hz, and control commands are sent with a delay kept between 18 and 30 milliseconds. The low-level interface operates at a frequency of 500 Hz, ensuring smooth real-time control. Communication between the control policy and the low-level interface uses LCM (Lightweight Communications and Marshalling). These frequencies describe physical-robot policy deployment, rather than world-model video-generation throughput.

\begin{itemize}
    \item \textbf{Marvin-WUJI.} Dual-Arm Dexterous Hand System. The first embodiment is built on the TIANJI M6 arms and WUJI dexterous hands. The hardware setup consists of dual Tianji 7-DOF robotic arms and dual Wuji 20-DOF dexterous hands, yielding a total of 54 degrees of freedom.
    \item \textbf{Astribot-S1.} Astribot-S1 Mobile Manipulator. The second embodiment utilizes the Astribot S1, an AI robot platform featuring a wheeled mobile base and dual 7-DOF robotic arms equipped with parallel grippers. Each arm provides 7 DOFs with a rated payload of 5\,kg (horizontal extension) and an end-effector repeatability of $\pm 0.1$\,mm. With a height of 170\,cm and an arm span of 194\,cm, Astribot-S1 provides a highly dynamic, mobile bimanual manipulation baseline compared to the stationary dexterous setup.
\end{itemize}


\noindent
\textbf{Real-world policy tasks.}
We evaluate downstream policy learning on four tasks that span varying levels of bimanual coordination, contact richness, and precision. These tasks are distributed across two distinct dual-arm robotic platforms: \textbf{Marvin-WUJI}, a stationary dual-arm robot equipped with dexterous hands, and \textbf{Astribot-S1}, a wheeled mobile dual-arm robot equipped with parallel grippers. The four tasks are defined as follows:

\textbf{Tasks performed on Marvin-WUJI (Dual-Arm Dexterous Hands):}
\begin{itemize}
    \item \textbf{(1) Plate-Placement:} The robot reaches toward a shelf, uses its dexterous hands to securely grasp a plate, and places it beside the oven on the tabletop, requiring precise spatial alignment and stable grasping of thin, fragile objects.
    \item \textbf{(2) Plate-Wiping:} A highly dexterous bimanual coordination task in which one hand grasps and stabilizes a plate while the other hand holds a cloth to wipe the plate's surface before returning the plate to the shelf, requiring coordinated bimanual manipulation.
\end{itemize}

\textbf{Tasks performed on Astribot-S1 (Wheeled Mobile Dual-Arm Grippers):}
\begin{itemize}
    \item \textbf{(3) Steak-Serving:} The robot approaches the dining area, uses its dual grippers to stably lift a plate containing a cooked steak, and accurately serves it onto the designated spot on the dining table, integrating mobile navigation with precise placement.
    \item \textbf{(4) Plate-Retrieval:} Following the meal, the robot identifies and grasps used, dirty plates left on the dining table, collects the associated cutlery, and returns the items to their designated positions.
\end{itemize}


\noindent
\textbf{Evaluation metrics.}
For world-model video prediction quality, we report PSNR, SSIM, and LPIPS, where higher PSNR/SSIM and lower LPIPS are better. Following DreamDojo~\citep{gao2026dreamdojo}, we evaluate open-loop rollouts conditioned on an initial frame and native robot-action sequences through the target-robot action encoder, autoregressively replacing the conditioning frame with the last predicted frame. At inference we apply classifier-free guidance on the action conditioning, $\hat{u}=u_{\varnothing}+1.15\,(u_{A}-u_{\varnothing})$, where $u_{A}$ is the velocity field conditioned on the action sequence and $u_{\varnothing}$ the same field with the sequence dropped, as defined in Sec.~\ref{sec:method}. For these quantitative benchmarks, the robot-action-conditioned model receives recorded control commands rather than latent actions extracted from future ground-truth frames. The qualitative action-following experiment instead uses source-video latent actions. These image-based metrics measure prediction fidelity rather than physical validity or action following in isolation. We therefore also examine qualitative action following under a fixed initial observation (Sec.~\ref{sec:action_following}). For policy learning, we report task success rate. For simulation policy benchmarks, we report average success rate under the official evaluation protocol. For real-world experiments, we report closed-loop success rate on the physical robots.


\begin{table}[t]
    \centering
    \caption{\textbf{Results on the DreamDojo GR-1 evaluation sets}~\citep{gao2026dreamdojo}. Baseline values are literature-reported measurements after robot post-training. }
    \label{tab:dreamdojo_eval}
    \small
    \setlength{\tabcolsep}{2pt}
    \begin{tabular}{lccccccccc}
        \toprule
        \multirow{2}{*}{Model} & \multicolumn{3}{c}{In-lab Eval} & \multicolumn{3}{c}{EgoDex Eval} & \multicolumn{3}{c}{DreamDojo-HV Eval} \\
        \cmidrule(lr){2-4} \cmidrule(lr){5-7} \cmidrule(lr){8-10}
        & PSNR$\uparrow$ & SSIM$\uparrow$ & LPIPS$\downarrow$ & PSNR$\uparrow$ & SSIM$\uparrow$ & LPIPS$\downarrow$ & PSNR$\uparrow$ & SSIM$\uparrow$ & LPIPS$\downarrow$ \\
        \midrule
        \rowcolor{gray!10}Cosmos-Predict2.5~\citep{ali2025world} & 20.58 & 0.77 & 0.22 & 19.95 & 0.79 & 0.22 & 18.27 & 0.75 & 0.24 \\
        DreamDojo-2B~\citep{gao2026dreamdojo}  & 21.11 & 0.77 & 0.22 & 20.41 & 0.78 & 0.23 & 18.81 & 0.75 & 0.24 \\
        \rowcolor{gray!10}DreamDojo-14B~\citep{gao2026dreamdojo} & 21.41 & 0.79 & 0.21 & 20.53 & 0.79 & 0.21 & 18.92 & 0.75 & 0.23 \\
        \rowcolor{tablepurple} \textbf{\our} & \textbf{21.42} & \textbf{0.82} & \textbf{0.20}  & \textbf{20.89} & \textbf{0.81} & \textbf{0.19} & \textbf{19.22} & \textbf{0.79}    & \textbf{0.21}    \\
        \bottomrule
    \end{tabular}
\end{table}


\begin{table}[t]
    \centering
    \caption{\textbf{Results and ablation study on our real-robot evaluation sets.}
    `N/A' denotes baseline implementations without support for Astribot-S1 mobile-base motion. PSNR, SSIM, and LPIPS are interpreted jointly as measures of video prediction fidelity.}
    \label{tab:rynnworld_eval}
    \small
    \setlength{\tabcolsep}{5pt}
    \begin{tabular}{lcccccc}
        \toprule
        \multirow{2}{*}{Method} & \multicolumn{3}{c}{Marvin-WUJI Eval} & \multicolumn{3}{c}{Astribot-S1 Eval} \\
        \cmidrule(lr){2-4} \cmidrule(lr){5-7}
        & PSNR$\uparrow$ & SSIM$\uparrow$ & LPIPS$\downarrow$ & PSNR$\uparrow$ & SSIM$\uparrow$ & LPIPS$\downarrow$ \\
        \midrule
        \rowcolor{gray!10}Cosmos3-Edge~\citep{agarwal2026cosmos} & 20.03 & 0.78 & 0.31 & 13.99 & 0.42 & 0.54 \\
        Cosmos3-Nano~\citep{agarwal2026cosmos} & 20.09 & \textbf{0.80} & 0.26 & 14.22  & 0.53 & 0.45 \\
        \rowcolor{gray!10}RynnWorld-Teleop~\citep{zhao2026rynnworldteleop} & 20.89 & 0.76 & 0.19 & N/A & N/A & N/A     \\
        \rowcolor{tablepurple} \textbf{\our} & \textbf{21.86} & \textbf{0.80} & \textbf{0.12} & \textbf{15.96} & \textbf{0.60} & \textbf{0.34}     \\
        \midrule

        \multicolumn{7}{c}{\textit{Ablation Study}} \\
        \rowcolor{gray!10}\our text-conditioned & 19.80 & 0.74 & 0.30 & 13.60 & 0.40 & 0.55 \\
        \our w/o latent-action pretraining & 21.32 & 0.78 & 0.16 & 15.31 & 0.55 & 0.39 \\
        \rowcolor{gray!10}\our w/o frame-aligned injection  & 21.14 & 0.76 & 0.18 & 15.08 & 0.54 & 0.41 \\
        \our w/o local modulation & 21.62 & 0.77 & 0.14 & 15.71 & 0.58 & 0.36 \\
        \bottomrule
    \end{tabular}
\end{table}


\subsection{Action-Conditioned World-Model Benchmarks}
\label{sec:benchmark_results}

We first evaluate \our as an action-conditioned world model on both public benchmarks and our own real-robot benchmarks.

\noindent
\textbf{DreamDojo benchmarks.} 
We consider three evaluation sets introduced by DreamDojo~\citep{gao2026dreamdojo}: \textbf{In-lab Eval}, \textbf{EgoDex Eval}, and \textbf{DreamDojo-HV Eval}. These sets probe whether human-video pretraining transfers to object interactions unseen in the robot post-training data.

\noindent
\textbf{\our real-robot benchmarks.}
To test simulation fidelity on our own hardware, we construct evaluation sets on both robotic platforms by recording ground-truth rollouts across a diverse range of bimanual manipulation scenarios and asking the world model to reproduce them from the same initial frame and action sequence. The evaluation episodes cover varied objects, scenes, and contact patterns, spanning both the stationary dexterous setup (\textbf{Marvin-WUJI}) and the mobile gripper setup (\textbf{Astribot-S1}), and use an episode-level split: for each task, we randomly reserve $5\%$ of the recorded episodes for testing and use the remaining $95\%$ for target-robot post-training. Episodes are split before temporal chunking so that chunks from the same demonstration do not appear in both sets. This task-stratified split evaluates held-out demonstrations rather than held-out tasks.

\noindent
\textbf{Results.}
Tab.~\ref{tab:dreamdojo_eval} reports predictive quality on the three GR-1 evaluation sets. \our achieves better reported PSNR, SSIM, and LPIPS than the listed baselines on all three evaluation sets. For example, LPIPS decreases from $0.21$ to $0.19$ relative to DreamDojo-14B on EgoDex Eval. These results support improved perceptual prediction fidelity. Since baseline measurements are taken from the literature, this comparison concerns complete systems and does not isolate the contribution of the latent-action interface from differences in backbone, data, or training.

Tab.~\ref{tab:rynnworld_eval} reports video prediction quality on our dual-arm platforms. These evaluation sets cover dexterous-hand manipulation and mobile bimanual coordination, extending the evaluation beyond the GR-1 benchmark. \our improves PSNR and LPIPS over the listed baselines on both platforms after target-robot adaptation, and ties the best SSIM on Marvin-WUJI. These results demonstrate applicability to stationary dexterous and mobile gripper embodiments; they do not establish zero-shot transfer to an unadapted robot. Ablations of the conditioning interface and the injection architecture are discussed in Sec.~\ref{sec:ablation}.

\begin{table}[t]
    \centering
    \caption{\textbf{Closed-loop success rates on real-world manipulation tasks.} 
    The upper block evaluates policies trained with digital-teleoperation data. 
    The lower block evaluates policies trained with large-scale synthetic mid-training. 
    Each applicable policy setting is evaluated with 35 trials per task on the designated robot. Success rates are percentages; bold values indicate the best result per task within each table block. `N/A' denotes baseline implementations without support for Astribot-S1 mobile-base motion. The Real only rows are the reference settings for both blocks.}
    \label{tab:real_world_all}
    \small
    \setlength{\tabcolsep}{2.2pt}
    \begin{adjustbox}{max width=\columnwidth}
    \begin{tabular}{ll cccc}
        \toprule
        \multirow{2}{*}{Method} & \multirow{2}{*}{Training Setting} 
        & \multicolumn{2}{c}{\textbf{Marvin-WUJI}} 
        & \multicolumn{2}{c}{\textbf{Astribot-S1}} \\
        \cmidrule(lr){3-4} \cmidrule(lr){5-6}
        & & \shortstack{Plate \\ Placement} 
        & \shortstack{Plate \\ Wiping} 
        & \shortstack{Steak \\ Serving} 
        & \shortstack{Plate \\ Retrieval} \\
        \midrule
        \multicolumn{6}{c}{\textit{Digital teleoperation}} \\
        \rowcolor{gray!10}$\pi_{0}$~\citep{black2024pi_0} & Real only & 80.00 & 71.43 & 37.14 & 45.71 \\
        \rowcolor{tablepurple} $\pi_{0}$ & Real + Digital Teleop & 88.57 & \textbf{94.29} & 51.43 & 51.43 \\
        
        \rowcolor{gray!10}$\pi_{0.5}$~\citep{intelligence2025pi05} & Real only & 82.86 & 77.14 & 40.00 & 45.71 \\
        $\pi_{0.5}$ & Digital Teleop only & 57.14 & 54.29 & 28.57 & 31.43 \\
        \rowcolor{gray!10}$\pi_{0.5}$ & Real + RynnWorld-Teleop~\citep{zhao2026rynnworldteleop} & 85.71 & 82.86 & N/A  & N/A  \\
        \rowcolor{tablepurple}$\pi_{0.5}$ & Real + Digital Teleop & \textbf{91.43} & 91.43 & \textbf{54.29} & \textbf{57.14} \\

        \midrule
        \multicolumn{6}{c}{\textit{Large-scale synthetic mid-training}} \\
        \rowcolor{tablepurple} $\pi_{0}$ & \our-80M + Real & 85.71 & 82.86 & 48.57 & 54.29 \\
        $\pi_{0.5}$  & Real + Ego2Robot~\citep{wang2026ego2robot} & 88.57 & 91.43 & N/A & N/A \\
        \rowcolor{tablepurple} $\pi_{0.5}$ & \our-80M + Real & \textbf{94.29} & \textbf{97.14} & \textbf{60.00} & \textbf{68.57} \\
        \bottomrule
    \end{tabular}
    \end{adjustbox}
\end{table}


\subsection{Qualitative Action Following}
\label{sec:action_following}

\begin{figure}[htbp]
    \centering
    \includegraphics[width=\linewidth]{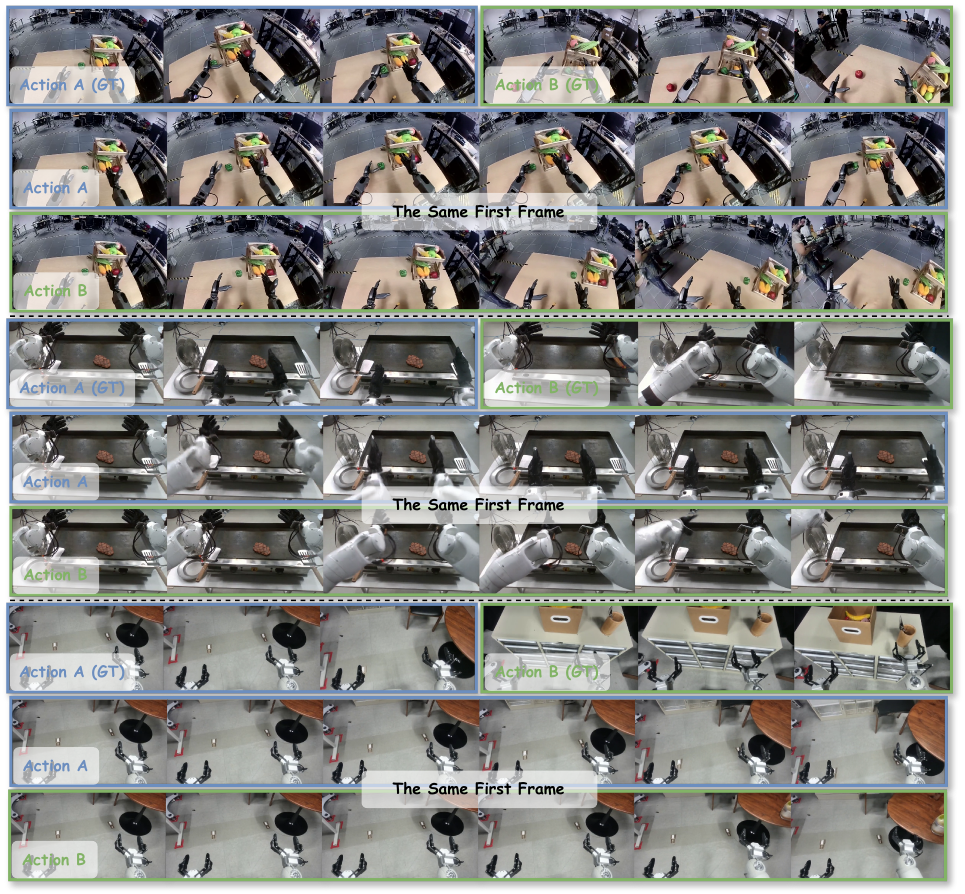}
    \caption{\textbf{Action following under a shared initial observation.}
    Each comparison uses latent-action sequences extracted from two different source videos. Both sequences are paired with the same initial frame to generate separate rollouts. Blue and green arrows associate source clips with their corresponding generated rollouts. The generated motion qualitatively follows the respective source sequence despite the shared initial scene.}
    \label{fig:action_following}
\end{figure}

We examine whether supplied latent actions guide the generated motion beyond the scene information provided by the first frame. For each comparison, we extract two latent-action sequences, $A^{(1)}$ and $A^{(2)}$, from different source videos and pair them with the same initial observation $I_0$. Encoding this observation as $c_0$, we generate
\begin{equation}
    \hat{V}^{(k)} \sim p_\theta(\cdot \mid c_0, A^{(k)}),
    \qquad k \in \{1,2\}.
\end{equation}
Fig.~\ref{fig:action_following} shows source clips and the corresponding generated rollouts. Although each pair shares an initial scene, its generated motion changes with the supplied action sequence and qualitatively resembles the corresponding source motion. These examples provide evidence of latent-action following under a shared initial observation. Unlike data editing, which changes the initial appearance while reusing source actions, this comparison holds the initial observation fixed and changes the action sequence. It is a qualitative assessment of visual transition control rather than a quantitative measure of physical accuracy or executable robot-action tracking.

\subsection{Digital-Teleoperation Data for Real-World Policy Learning}
\label{sec:digital_teleop}

We next test whether data collected through interactive digital teleoperation can train effective real-world policies.

\begin{figure}[htbp]
    \centering
    \includegraphics[width=\linewidth]{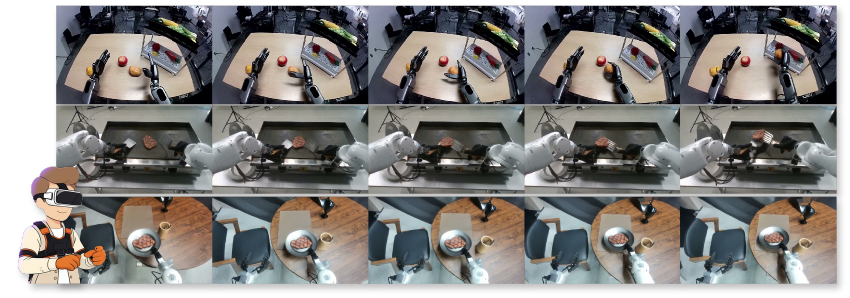}
    \caption{
    \textbf{Interactive digital teleoperation.}
    Operator control signals are retargeted to target-robot actions and condition \our together with a reference scene; the recorded actions and generated frames form synthetic demonstrations without physical robot execution.
    }
    \label{fig:digit_teleop}
\end{figure}

\noindent
\textbf{Data collection.}
An operator provides control signals through a teleoperation interface. These signals are retargeted to the target robot action space and used to condition \our. Given an initial conditioning frame and an action chunk, \our generates the corresponding future video latent sequence. The system records the synchronized robot actions and generated video frames, producing synthetic demonstrations without physical robot execution. Fig.~\ref{fig:digit_teleop} illustrates this loop. The digital-teleoperation episodes are synthetic data and are distinct from the physically collected demonstrations summarized in Appendix~\ref{app:real-data}.

\noindent
\textbf{Experimental settings.}
We compare \textbf{Real only} and \textbf{Real + Digital Teleop} for both $\pi_{0}$~\citep{black2024pi_0} and $\pi_{0.5}$~\citep{intelligence2025pi05}. The former uses real robot demonstrations, while the latter additionally uses generated digital-teleoperation episodes. For $\pi_{0.5}$, we also evaluate \textbf{Digital Teleop only}, using synthetic demonstrations without task-specific real-demonstration training.
We additionally compare against \textbf{Real + RynnWorld-Teleop}, which combines real demonstrations with synthetic teleoperation data generated by RynnWorld-Teleop~\citep{zhao2026rynnworldteleop}. Our RynnWorld-Teleop baseline implementation does not support Astribot-S1 mobile-base motion, so results are reported only on Marvin-WUJI. All policies are evaluated closed-loop on the real robot platforms.

\noindent
\textbf{Results.}
The upper block of Tab.~\ref{tab:real_world_all} reports closed-loop success rates on the real-world tasks. Adding digital-teleoperation data improves success rates on all four tasks for both backbones. For $\pi_{0.5}$, Plate-Placement improves from $82.86\%$ to $91.43\%$, and Steak-Serving from $40.00\%$ to $54.29\%$. The \textbf{Digital Teleop only} setting achieves nonzero success on all four tasks but remains below Real only on every task, supporting synthetic data as an augmentation rather than a complete replacement for real demonstrations in this setting. On Marvin-WUJI, the $\pi_{0.5}$ Real + Digital Teleop setting also exceeds Real + RynnWorld-Teleop by approximately $5.71$ and $8.57$ percentage points on Plate-Placement and Plate-Wiping, corresponding to two and three additional successful trials. Each reported rate is based on 35 trials; we interpret these as observed improvements rather than claims of statistical significance.

\subsection{Scalable Data Editing and Synthesis for Policy Learning}

Beyond interactive teleoperation, we use \our as a scalable data engine that converts large-scale human interaction videos into robot-action-labeled data.

\noindent
\textbf{Synthetic corpus.}
Given a source human video, we extract its latent-action sequence using the frozen \lam, edit the first frame to replace the human embodiment or original environment with the target robot embodiment or a robot-compatible scene, and condition \our on the edited frame and the original latent-action sequence to generate a robotized rollout. Robot commands are obtained by relabeling the source motion through retargeting and inverse kinematics; implementation details are provided in Appendix~\ref{app:ik-relabeling}. Fig.~\ref{fig:data_edit} illustrates this pipeline.

\noindent
\textbf{Large-scale synthesis corpus.}
We apply this pipeline to EPIC-KITCHENS~\citep{damen2020epic}, EgoDex~\citep{hoque2025egodex}, and Something-Something~\citep{goyal2017something}. After filtering for visual quality, temporal consistency, action feasibility, and interaction plausibility, we obtain approximately 84.8M frames, corresponding to roughly 800 hours of synthetic robot video-action data. We refer to this corpus as \our-80M. Tab.~\ref{tab:data_engine_sources} groups the generated frames by their source dataset.

\noindent
\textbf{Simulation policy benchmarks.}
We examine whether synthetic mid-training improves policy learning in simulation. Starting from pretrained $\pi_{0}$ and $\pi_{0.5}$ checkpoints, we perform mid-training on \our-80M and then evaluate on standard policy benchmarks. We use LIBERO~\citep{liu2023libero} and LIBERO-Plus. LIBERO evaluates language-conditioned manipulation, while LIBERO-Plus evaluates robustness under perturbations.

\begin{table}[t]
    \centering
    \caption{\textbf{Synthetic corpus by source dataset.} The \our data engine processes these human interaction videos into robot-action-labeled synthetic episodes. After filtering, the total corpus contains approximately 800 hours.}
    \label{tab:data_engine_sources}
    \small
    \begin{tabular}{lc}
        \toprule
        Source Dataset & Synthetic Frames \\
        \midrule
        \rowcolor{gray!10}EPIC-KITCHENS~\citep{damen2020epic} & 2.5M \\
        EgoDex~\citep{hoque2025egodex}  & 80.8M \\
        \rowcolor{gray!10}Something-Something~\citep{goyal2017something} & 1.5M \\
        \rowcolor{tablepurple} \textbf{\our-80M (total)} & 84.8M \\
        \bottomrule
    \end{tabular}
\end{table}

\noindent
\textbf{Real-world policy learning.}
We further test whether synthetic mid-training improves real-world policy learning. For each policy backbone, we compare two settings:
(i) \textbf{Real post-training only}, where the policy is fine-tuned solely on real robot demonstrations;
and (ii) \textbf{\our-80M mid-training + Real post-training}, where the policy is first mid-trained on the synthetic corpus and then fine-tuned on the same real robot demonstrations.
We additionally compare with \textbf{Real + Ego2Robot}, which combines real demonstrations with synthetic data generated by Ego2Robot~\citep{wang2026ego2robot}. Our Ego2Robot baseline implementation does not support Astribot-S1 mobile-base motion, so results are reported only on Marvin-WUJI. All policies are evaluated closed-loop on the real robot platforms.

\noindent
\textbf{Results.}
Tab.~\ref{tab:exp_libero} reports simulation benchmark results after synthetic mid-training. Synthetic mid-training improves $\pi_{0}$ from $94.4\%$ to $96.1\%$ on LIBERO and from $53.6\%$ to $79.4\%$ on LIBERO-Plus. For $\pi_{0.5}$, the corresponding changes are $96.9\%$ to $97.8\%$ and $84.4\%$ to $84.8\%$; the observed gains are therefore larger for $\pi_{0}$ than for $\pi_{0.5}$. The lower block of Tab.~\ref{tab:real_world_all} reports improvements over Real only on all four real-world tasks for both backbones. For $\pi_{0.5}$, Steak-Serving rises from $40.00\%$ to $60.00\%$, and Plate-Retrieval from $45.71\%$ to $68.57\%$. On Marvin-WUJI, $\pi_{0.5}$ with \our-80M + Real also exceeds Real + Ego2Robot by approximately $5.71$ percentage points on each task, corresponding to two additional successful trials. These comparisons support the utility of the synthetic corpus in the evaluated training pipelines. They do not isolate generation quality from differences in synthetic-data quantity or additional training compute.


\begin{table}[t]
  \caption{\textbf{Policy success rates on LIBERO~\citep{liu2023libero} and LIBERO-Plus.} Values are percentages. Each synthetic-data variant uses \our-80M for mid-training before downstream evaluation.}
  \label{tab:exp_libero}
  \small
  \centering
  \begin{tabular}{l c c}
    \toprule
    \textbf{Method} & \textbf{LIBERO} & \textbf{LIBERO-Plus} \\
    \midrule
    $\pi_{0}$~\citep{black2024pi_0}  & 94.4 & 53.6  \\
    \rowcolor{tablepurple} \textbf{$\pi_{0}$ + \our-80M} & \textbf{96.1} & \textbf{79.4}\\
    \hline
    $\pi_{0.5}$~\citep{intelligence2025pi05}   & 96.9 & 84.4 \\
    \rowcolor{tablepurple} \textbf{$\pi_{0.5}$ + \our-80M} & \textbf{97.8} & \textbf{84.8} \\
    \bottomrule
  \end{tabular}
\end{table}


\begin{figure}[htbp]
    \centering
    \includegraphics[width=\linewidth]{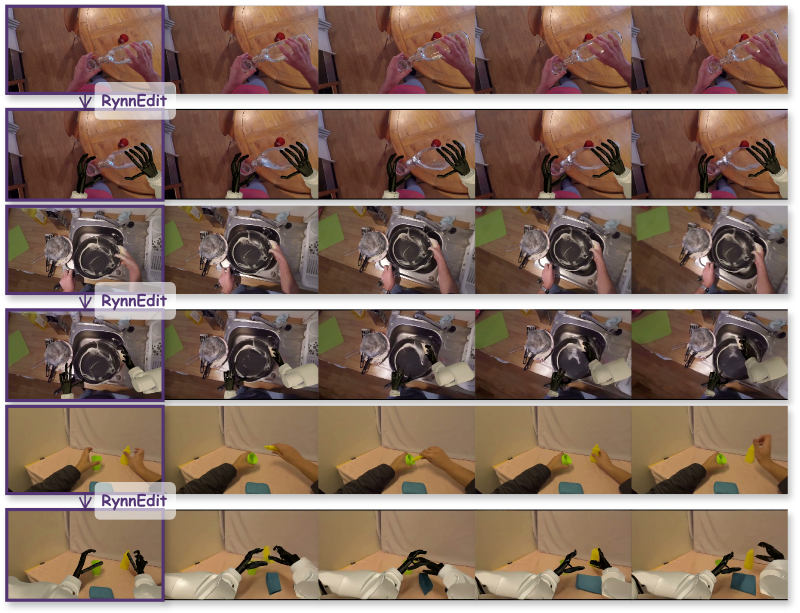}
    \caption{
    \textbf{Scalable data editing.}
    Given a source video, we edit its first frame to replace the embodiment or environment, then condition \our on the edited frame and the source latent-action sequence to synthesize a robotized rollout.
    }
    \label{fig:data_edit}
\end{figure}

\subsection{Ablation Studies}
\label{sec:ablation}

\paragraph{Action-conditioning interface.}
Tab.~\ref{tab:rynnworld_eval} compares conditioning interfaces on the real-robot evaluation sets. Replacing the action-conditioning interface with task-level text reduces PSNR from $21.86$ to $19.80$ on Marvin-WUJI and from $15.96$ to $13.60$ on Astribot-S1; LPIPS increases from $0.12$ to $0.30$ and from $0.34$ to $0.55$, respectively. SSIM also decreases on both platforms. These results are consistent with frame-level action information providing a more specific prediction signal than task-level language.
The variant without latent-action pretraining performs worse than the full model on all reported video prediction metrics
Native robot commands are also used to evaluate the full model after target-robot adaptation; the latent interface serves its cross-embodiment pretraining stage. The latent interface additionally allows videos without native robot controls to contribute to pretraining; the present ablation does not separately identify how much of the observed gain comes from representation design versus access to additional video data.

\paragraph{Conditioning architecture.}
Removing frame-aligned injection while retaining global action tokens decreases PSNR by $0.72$ on Marvin-WUJI and $0.88$ on Astribot-S1. SSIM decreases from $0.80$ to $0.76$ and from $0.60$ to $0.54$, while LPIPS increases from $0.12$ to $0.18$ and from $0.34$ to $0.41$. Removing only local feature modulation produces smaller degradations: PSNR decreases by $0.24$ and $0.25$, respectively. The full model achieves the best metrics among these variants on both platforms. These results support combining global action tokens with local frame-aligned conditioning, while indicating that the additive pathway retains useful conditioning when modulation is removed. Image-level metrics alone do not determine whether the errors arise from contact timing, camera motion, or other visual changes.

%% file: Sec/5-conclusion.tex
\section{Conclusion}
We presented \our, a cross-embodiment video world model that uses shared continuous latent actions as a conditioning interface for learning visual dynamics from heterogeneous interaction videos. Building on the 3D-grounded \lam representation, the model combines global action tokens with temporally aligned local conditioning to connect motion descriptors with future video latents. This enables world-model pretraining across human, robot, and simulation videos without requiring their native action spaces to be aligned, followed by target-robot adaptation through an embodiment-specific action encoder.

Experiments show strong video prediction fidelity on public benchmarks and two real-robot platforms, while qualitative comparisons demonstrate that generated motions respond to different supplied latent-action sequences under the same initial observation. The model also supports approximate cross-embodiment motion transfer and native-action-conditioned digital teleoperation. Synthetic data generated through both routes improves downstream policy success in the evaluated settings. Together, these findings support shared latent actions as a practical interface for cross-embodiment world-model learning and controllable robot-data generation.

\paragraph{Limitations.}
Latent actions are inferred from observed visual transitions and may encode camera motion or appearance information as well as manipulation. Qualitative action following and image-based prediction metrics therefore do not establish causal identification or physical correctness. In the synthesis pipeline, retargeted robot commands and generated observations are obtained through separate processes; temporal alignment and filtering cannot guarantee their physical consistency. The reported real-world success rates use 35 trials per task, and small differences should be interpreted accordingly. Moreover, synthetic mid-training adds both data and optimization compute, whose contributions are not separated by the current comparisons. Future work will address long-horizon consistency, action-observation alignment, and broader transfer across embodiments.

%% file: Sec/Appendix.tex
\section{Teleoperation Setup}
\begin{figure}
  \centering
    \includegraphics[width=0.9\linewidth]{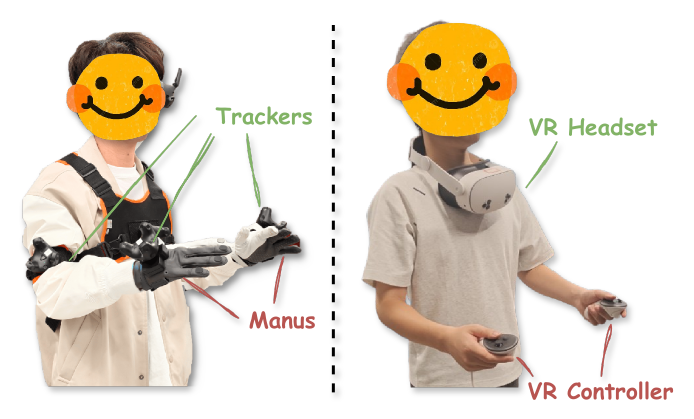}\\
  \caption{\textbf{Marvin-WUJI demonstration collection.} An operator controls dual robot arms and dexterous hands using motion trackers and data gloves.}
  \label{fig:operator}
\end{figure}

For Marvin-WUJI, we collect real-world demonstrations through the teleoperation system shown in Fig.~\ref{fig:operator}. The hardware setup consists of dual Tianji 7-DOF robotic arms and dual Wuji 20-DOF dexterous hands, yielding a total of 54 degrees of freedom.                            

For arm control, the human operator wears five HTC Vive trackers (mounted on the chest, both wrists, and both upper arms). The system computes wrist-to-chest relative transforms at 100-120 Hz and feeds them into a Pinocchio-based inverse kinematics solver running in a separate process. The resulting joint commands are further smoothed by a Ruckig trajectory generator with velocity, acceleration, and jerk constraints before being sent to the robot arms at 200 Hz.                  

For hand control, the operator wears Manus data gloves. The raw glove signals are converted to a 21-point MediaPipe hand skeleton format and retargeted to the 20-DOF Wuji hand joint space via a dedicated retargeting module, with an exponential moving average filter applied for motion smoothing.

\section{Real-World Demonstration Corpus}
\label{app:real-data}

The physically collected demonstration corpus is the same as that used in RynnVLA-Latent~\citep{zhao2026rynnvlalatent}. It contains 3,850 episodes across six manipulation tasks on the real-robot platforms described in Sec.~\ref{sec:experiments}. Tab.~\ref{tab:teleop_data_stats} reports the collected episode counts. These counts describe the full corpus, not the number of episodes in an individual post-training split or the amount of generated digital-teleoperation data. For world-model adaptation and evaluation, we randomly split episodes within each task into $95\%$ for post-training and $5\%$ for testing. The split is defined at the episode level before video chunking. Thus all chunks from a held-out episode remain outside the post-training split, and the test tasks themselves are represented in training. The downstream policy experiments in this paper evaluate four of the six collected tasks.

\begin{table}[t]
\centering
\caption{\textbf{Collected real-world demonstrations.} The corpus is shared with RynnVLA-Latent. Counts refer to physically collected episodes, not synthetic data.}
\label{tab:teleop_data_stats}
\small
\begin{tabular}{lp{0.62\linewidth}r}
\toprule
\textbf{Task} & \textbf{Description} & \textbf{Episodes} \\
\midrule
\rowcolor{gray!10}Plate-Placement & Take a plate from a shelf and place it beside the oven. & 500 \\
Steak-Cooking & Flip a steak in a pan. & 800 \\
\rowcolor{gray!10}Plate-Wiping & Take a dirty plate from a shelf, wipe it with a cloth, and return it. & 500 \\
Table-Cleaning & Collect scattered items into a box and place the box on a shelf. & 850 \\
\rowcolor{gray!10}Steak-Serving & Carry a steak plate to the dining area and place it on a placemat. & 600 \\
Plate-Retrieval & Collect a dirty plate and cutlery and return them to their designated positions. & 600 \\
\rowcolor{tablepurple} \textbf{Total} & & \textbf{3,850} \\
\bottomrule
\end{tabular}
\end{table}

We use the world-model chunking protocol from Sec.~\ref{sec:method}: \lam descriptors are extracted at a four-frame stride, with up to 20 latent-action transitions in an 81-frame video chunk. This differs from the policy-specific action-chunk protocol in the companion VLA work. Sharing the demonstration corpus does not imply identical training objectives or temporal chunk lengths.

\section{Details of Pseudo Robot Action Relabeling}
\label{app:ik-relabeling}

For source videos without target robot actions, we estimate the relevant hand or end-effector trajectory from the source motion. The estimated trajectory is retargeted to the target robot workspace through coordinate mapping, workspace scaling, and embodiment-specific retargeting. We then solve inverse kinematics to obtain target robot joint commands. During IK solving, we enforce joint limits, velocity limits, and temporal smoothness. The resulting actions are treated as pseudo-labels rather than ground-truth robot commands. These pseudo-labels are subsequently filtered together with the generated videos before being used for policy training.


\section{Baseline Implementation Details} 
\noindent 
\textbf{Ego2Robot.} We reproduce Ego2Robot~\citep{wang2026ego2robot} in our pipeline. Specifically, we audit each raw video and its corresponding HaWoR~\citep{zhang2025hawor} motion-capture trajectory, geometrically retarget the HaWoR hand poses to the 20-degree-of-freedom (DoF) Wuji hand, and resample the trajectories using zero-order hold or interpolation. At the source resolution, we use the projected HaWoR keypoints as prompts for SAM3~\citep{carion2026sam} to segment both hands and refine the segmentation using text prompts. We then take the union of the left- and right-hand masks, dilate it by 4 pixels, and remove the hands using a video object removal model, such as MiniMax-Remover~\citep{zi2026minimax} or ProPainter~\citep{zhou2023propainter}. In parallel, on the CPU, we estimate the robot base pose through a coarse-to-fine inverse kinematics (IK) search in MuJoCo, evaluating 81 candidates in the coarse stage. The visual strategy is decoupled from the segmentation masks, and this branch outputs \texttt{qpos54} and an IK report. Finally, we render the TIANJI M6 arm and Wuji hand using MuJoCo's EGL-based off-screen rasterization and composite the rendered robot onto the inpainted video using an end-effector-locked projection strategy.

We further introduce quality gates for sample selection. A sample is marked as suitable for training only if it passes both the inpainting content quality checks and the strict IK and projection thresholds. The inpainting checks require a peak signal-to-noise ratio (PSNR) outside the removal mask of at least $30\,\mathrm{dB}$, relaxed to $22\,\mathrm{dB}$ for videos with camera motion; the absence of entirely black frames; and successful checks of per-frame spatial standard deviation and temporal response. The rendering criteria require a position error of at most $5\,\mathrm{mm}$, a rotation error of at most $5^\circ$, an end-effector projection error of at most $3$ pixels, a 95th-percentile retargeting landmark error of at most $0.1$, and a finger-direction consistency score of at least $0.9$.
